\documentclass[10pt]{article}
\usepackage{spconf,amsmath,graphicx}

\usepackage{cite}
\usepackage{amsmath,amssymb,amsfonts}
\usepackage{algorithmic}
\usepackage{textcomp}
\usepackage{xcolor}
\usepackage{subfigure}

\usepackage{enumitem}
\setlist{nosep, leftmargin=14pt}

\title{RadHop-Net: A lightweight radiomics-to-error regression for false positive reduction in MRI prostate cancer detection}
\name{Vasileios Magoulianitis, Jiaxin Yang, Catherine A. Alexander, C.-C. Jay Kuo}
\address{Ming Hsieh Department of Electrical and Computer Engineering \\ University of Southern California, Los Angeles, CA, USA}
\begin{document}
%\ninept
%
\maketitle

\begin{abstract}
Clinically significant prostate cancer (csPCa) is a leading cause of cancer death in men, yet it has a high survival rate if diagnosed early. Bi-parametric MRI (bpMRI) reading has become a prominent screening test for csPCa. However, this process has a high false positive (FP) rate, incurring higher diagnostic costs and patient discomfort. This paper introduces RadHop-Net, a novel and lightweight CNN for FP reduction. The pipeline consists of two stages: Stage 1 employs data-driven radiomics to extract candidate ROIs. In contrast, Stage 2 expands the receptive field about each ROI using RadHop-Net to compensate for the predicted error from Stage 1. Moreover, a novel loss function for regression problems is introduced to balance the influence between FPs and true positives (TPs). RadHop-Net is trained in a radiomics-to-error manner, thus decoupling from the common voxel-to-label approach. The proposed Stage 2 improves the average precision (AP) in lesion detection from $0.407$ to $0.468$ in the publicly available pi-cai dataset, also maintaining a significantly smaller model size than the state-of-the-art.
\end{abstract}

\section{Introduction}
Prostate cancer (PCa) is one of the leading occurring cancers in men, yet if diagnosed early, the five year survival rate is almost 100\%~\cite{siegelstats}. In clinical practice, the prostate-specific antigen (PSA) biomarker test has been used for years to screen patients for clinically significant PCa (csPCa) and determine those who need a biopsy. However, this procedure has a reportedly high false positive rate~\cite{galey_psa}, thus increasing the number of unnecessary biopsies. Bi-parametric magnetic resonance imaging (bpMRI) examination is a primary screening for csPCa to determine which patients need to undergo a biopsy. Although, the MRI reading entails expertise and is subjective to the reader. Hence, there is an inevitably high discordance rate~\cite{seo_pirads} among radiologists. 

Computer-aided diagnosis (CAD) tools powered with modern AI algorithms can expedite the MRI readout, provide more objectivity in decision-making, and increase the sensitivity and specificity of the csPCa detection. Several works have been published, either using traditional methods by employing handcrafted radiomics~\cite{algohary_radiomics, filos_radiomics}, or deep learning models~\cite{Vente_Unet, Huang_Unet, isensee2021nnu}, with certain works proposing attention mechanisms ~\cite{Zhang_crossmodal, Saha_fpreduction} to enhance the feature learning. Recently, a novel and lightweight data-driven radiomics extraction model was proposed, named RadHop~\cite{magoul_RadHop}, which demonstrated a competitive performance. RadHop is based on the Saab transform~\cite{Saab}, which provides a more transparent feature extraction framework.  

\begin{figure*}[t]
\centerline{\includegraphics[width=1.0\linewidth]{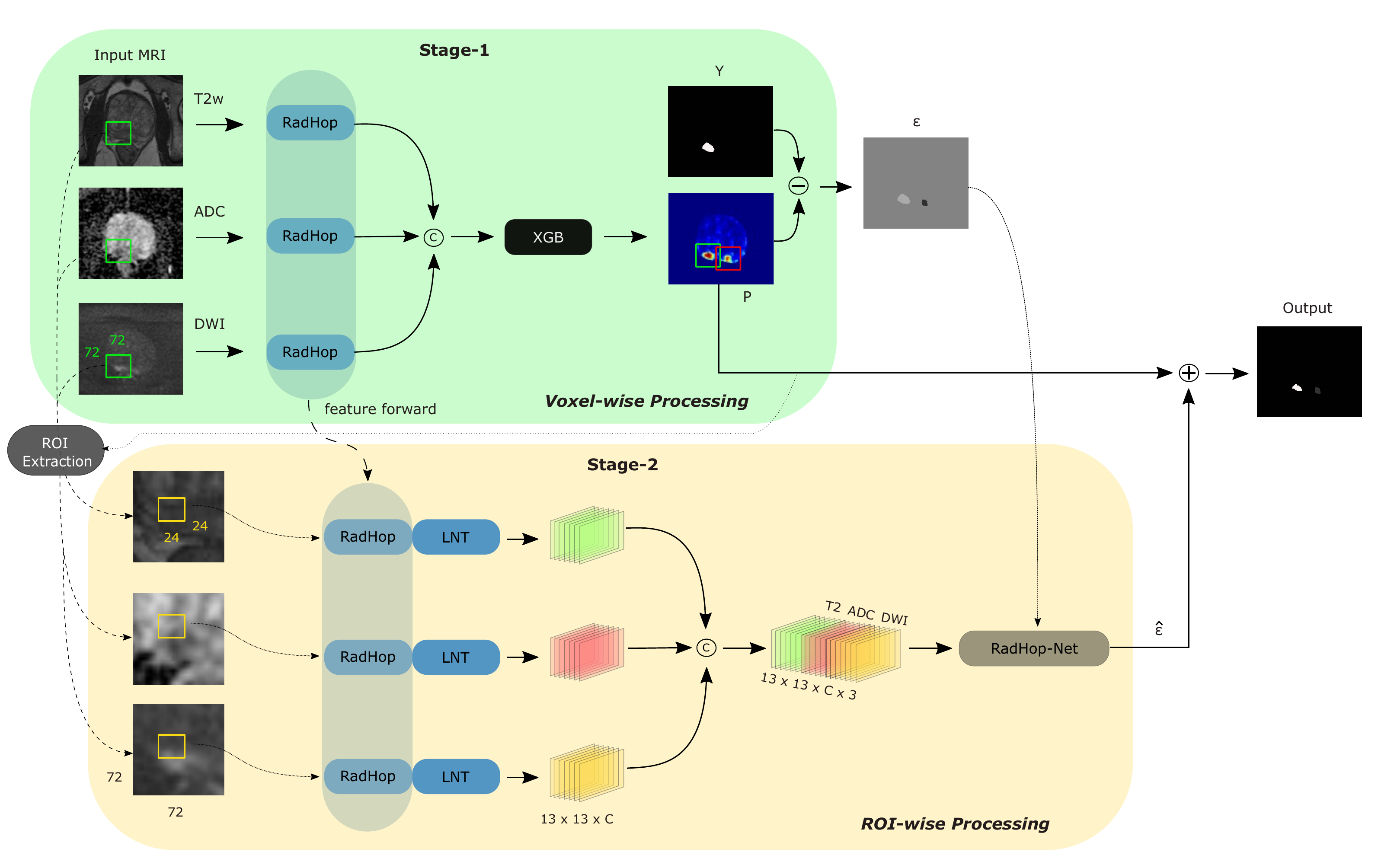}}
\vspace{-4mm}
\caption{The proposed pipeline has two stages of processing. Stage 1 extracts RadHop features from each sequence to predict a heatmap and extract candidate ROIs. In Stage 2, RadHop-Net expands the receptive field about each ROI and compensates for the probability error from Stage 1 predictions.}
\label{fig:pipeline}
\end{figure*}

One challenge with CAD algorithms for csPCa detection from MRI is the high false positive (FP) rate at the lesion level, which in turn, can give inaccurate predictions at the patient level and also mislead the MRI-based targeted biopsy. In literature, certain works~\cite{Yu_contextual, Saha_fpreduction} have shown that increasing the contextual information about the suspicious regions of interest (ROIs) enhances the feature representation and helps to discern the FP over the TPs ROIs. This work builds on this idea, introducing RadHop-Net, a custom convolutional neural network (CNN), to consider a larger area about each ROI and reduce the FP rate.

The overall pipeline includes two stages: Stage 1 uses RadHop radiomics to detect candidate ROIs that are suspicious for harboring csPCa. In Stage 2, RadHop-Net is used to ``correct" the Stage 1 probability for each ROI, by increasing the receptive field and contextual information about each ROI. Toward this effort, there are three novelties: (1) the proposed RadHop-Net is trained on the RadHop feature space to replace the early CNN layers and provide a lightweight model size. (2) RadHop-Net is trained as a regression model to compensate for the probability error (residue) from Stage 1. (3) since the FPs used to produce the regression values are in practice many more than the TPs, a novel loss function tailored to this problem is introduced to mitigate the effect of unbalanced regression values. The overall pipeline is devised to provide a competitive detection performance,  with a significantly smaller model size. 

\begin{figure*}[t]
\centerline{\includegraphics[width=1.0\linewidth]{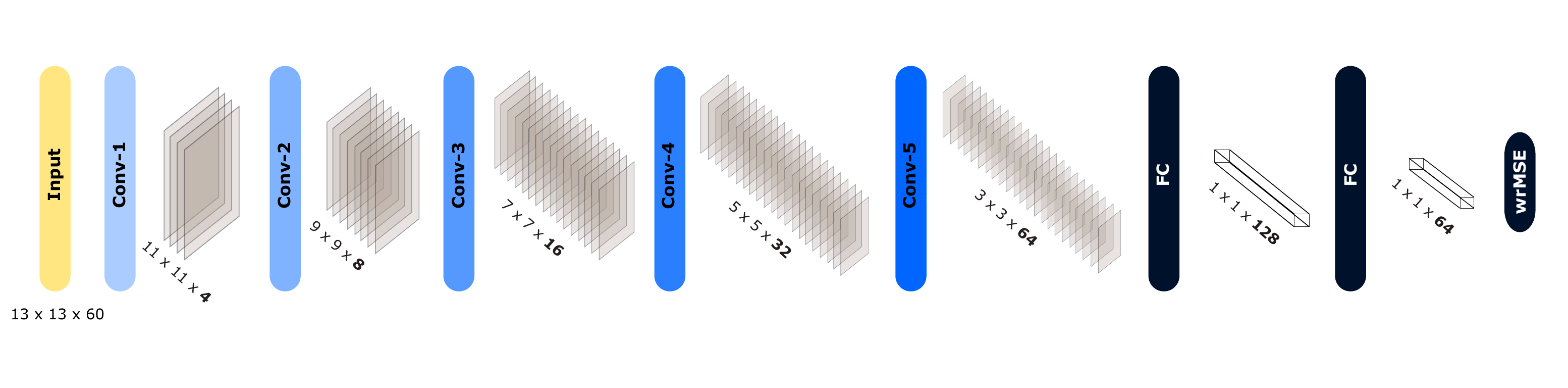}}
\vspace{-10mm}
\caption{The architecture of the proposed RadHop-Net for Stage 1 probability error prediction.}
\label{fig:modelNN}
\end{figure*}

\vspace{-1mm}

\section{Methodology} \label{sec:methodology}

The proposed pipeline adopts a two-stage approach, where Stage-1 (Section~\ref{subsec:stage1}) extracts the suspicious ROIs with a probability of being csPCa, while Stage-2 (Section~\ref{subsec:stage2}) performs FP reduction using the RadHop-Net model (Fig.~\ref{fig:pipeline}).

\vspace{-2mm}

\subsection{Stage 1 -- Candidate ROIs Detection} \label{subsec:stage1}

 To scan the prostate for csPCa lesions, the RadHop feature extraction module is applied independently to each input sequence to extract data-driven radiomics in a voxel-wise manner. The core operation in RadHop architecture~\cite{magoul_RadHop} is based on the linear Saab~\cite{Saab} transform, which decomposes the input signal into a rich spatial-spectral representation without any supervision. RadHop features are interpretable since they can be mapped back to the original domain (i.e., linearity property), increasing the overall pipeline's transparency. The premise behind RadHop is that csPCa has more information on certain spectral components, which can help classify local areas of the prostate. The optimal feature extraction window for RadHop was empirically found to be $24\times24$. Larger windows induce more noise into features, while smaller windows do not provide the necessary contextual information. It is worth noting that RadHop has a significantly smaller model size and complexity when compared with other state-of-the-art works, thereby enabling a low complexity lesion segmentation in Stage 1. 

After RadHop radiomics extraction, feature selection is applied to each sequence to filter out the less informative dimensions. From each sequence, we keep the $800$ most informative features~\cite{magoul_RadHop}. At the end of Stage 1, the RadHop radiomics from the three modalities are concatenated to train an Extreme Boosting Classifier (XGB) to predict heatmap $P$. The feature extraction and predictions are performed for each slice independently. For more details about RadHop and its hyperparameter settings, we refer the reader to~\cite{magoul_RadHop}. Finally, to yield candidate ROIs for Stage 2, the predicted heatmap $P$ is binarized using a relatively low probability threshold ($T_p=0.3$) to ensure a high sensitivity. The assigned probability of each ROI, $P_{roi}$, is its maximum voxel probability.

\vspace{-3mm}

\subsection{Stage 2 -- FP Reduction With RadHop-Net} \label{subsec:stage2}

One of the RadHop feature limitations is that their receptive field is relatively small, to differentiate between TPs and ``hard" FPs. The goal in Stage 2 is to enhance the feature space by expanding the receptive field about each ROI, including more contextual information. As such, the ultimate objective is to increase the probability of the TP ROIs and diminish that of the FPs. Since there is a prior csPCa probability for each ROI from Stage 1, we view this problem as regression and try to ``correct" the error residue $\varepsilon$ of the probability predictions between the extracted ROIs from the heatmap $P_{roi}$ and the ground truth annotations $Y_{roi}$:
\begin{equation}\label{eq:residue}
   \varepsilon_{roi} = Y_{roi} - P_{roi}, \hspace{7mm} Y \in \{0,1\}, \hspace{2mm} P \in [0,1].
\end{equation}

That is, TPs will produce non-negative values, while FPs will produce non-positive values.

\subsubsection{Radiomics-to-Error Prediction}

Since the receptive field expansion we consider in Stage 2 is $72\times72$, a CNN model is a natural choice for learning the complex neighborhood dynamics of the RadHop feature space between TPs and FPs. To this end, we propose RadHop-Net, a custom CNN model that receives an input patch centered at an ROI detected from Stage 1. Although most models in the literature are trained voxel-to-label, RadHop-Net is trained in a radiomics-to-error manner, using the RadHop features as input. One benefit is that during the designing of the model architecture, we can skip the early CNN layers --reducing the complexity--, since the RadHop provide a discriminant features representation that covers a $24\times 24$ area already. The low-scale information is retained from the RadHop features, while RadHop-Net helps in learning the large-scale (contextual) information.

\vspace{-2mm}

\subsubsection{RadHop-Net Architecture}

The input layer has a size of $13\times13\times C\times3$, including patches of size $24\times24$ extracted with a stride $4$ from the $72\times72$ area. $C$ is the number of RadHop features from each sequence, already extracted in Stage 1 and forwarded to Stage 2. Since we have three input modalities, the channel dimension of each tensor is $C\times3$. However, the RadHop feature dimension is relatively large to be used as input to a CNN and may cause overfitting. The Linear Normal Transform (LNT)~\cite{Wang_LNT} is used to extract more discriminant features using linear regression on random subsets of the RadHop features and also achieve dimensionality reduction. It helps provide a more compact RadHop feature representation that reduces the model complexity. We employ LNT to extract $20$ features from each modality (i.e., $C=20$), each resulting from a subset of $200$ RadHop features. 

To expand the receptive field using the CNN, we use $5$ successive convolutional layers --no max pooling in between-- and valid padding to increase the receptive field only through convolutions. At the end, two fully connected (FC) layers are placed before the final linear layer. The RadHop-Net architecture details are illustrated in Fig.~\ref{fig:modelNN}.

\vspace{-2mm}

\subsubsection{Weighted Residue MSE (wrMSE) Loss}
For regression problems, the Mean Square Error (MSE) is a popular choice as a loss function. However, in the csPCa detection problem, FPs candidates from stage 1 outnumber significantly the TP ones. Therefore, during training, the gradients of the negative values will dominate the weight updates, shifting the regression toward the negative values. To mitigate this issue and balance the RadHop-Net training, we introduce the weighted residue MSE (wrMSE) loss to weight the error according to the regressed value exponentially (see Eq.~\ref{eq:Loss}). Before RadHop-Net, the regression values are linearly mapped from the $(-1,1)$ range to $(0,1)$. A hyperparameter gamma $\gamma$ is used to control the weighting effect. The wrMSE loss for $N$ ROIs is calculated as:

\vspace{-4mm}

\begin{equation} \label{eq:Loss}
L_{wrMSE} = -\frac{1}{N} \sum_{i=1}^{N} \varepsilon_{roi}(i)^2 \cdot (\frac{1}{log(Y(i)_{roi})})^\gamma
\end{equation}

In inference mode, RadHop-Net predicts a residue correction $\hat{\varepsilon}$ for each ROI and adds it up to the original Stage 1 ROI probability. It yields the final output prediction (hard blobs with an assigned probability), where we measure the overall pipeline's patient and lesion level performance. For the patient-level probability, we consider the maximum probability among all the detected ROIs after Stage 2 updates.

\section{Experimental Results} \label{sec:Experimental}

\subsubsection{Dataset and Pre-processing}

To validate the lesion detection performance of the proposed pipeline and the effectiveness of RadHop-Net for FP reduction, we use the data from the PI-CAI Challenge~\cite{picai}. The patient cohort includes bpMRI data of $1,500$ patients, and a $5$-fold cross-validation scheme was adopted. The prostate gland masks are predicted from the nnU-Net model~\cite{isensee2021nnu}.

For data pre-processing, all the input sequences and annotation masks are registered on the T2w and further resampled to a voxel spacing of $3mm \times 0.25mm\times 0.25mm$ to standardize the MRIs from different scanners and clinical centers. All image values are normalized to $[0,1]$ range using the $0.05$ and $99.5$ percentiles of their intensity histogram.

\subsection{RadHop-Net Training}

To train the RadHop-Net, we use a batch size of $4096$ ROIs using the RMSProp optimizer with a learning rate of $10^{-4}$. For data augmentation, we apply horizontal and vertical flipping, as well as random shift changes in $[-5,5]$ and rotation in $[-10^{\circ},10^{\circ}]$. The $\gamma$ hyperparameter of the wrMSE loss function was empirically set at $0.95$. We train the model for $20$ epochs monitoring the validation Area Under the Curve (AUC) performance of the model after compensating the predicted residue error $\hat{\varepsilon}$ on the ROI probability from Stage 1 to evaluate the detection performance. 

\subsection{Results \& Comparisons}

\vspace{-5mm}

\begin{table}[htbp]
\caption{Performance benchmarking on the pi-cai dataset. AUROC measures the patient level and AP the lesion level performance.}
\begin{center}
\begin{tabular}{c c c}
\hline
   \textbf{Method / Setting}  & \textbf{AUROC} & \textbf{AP}  \\
\hline
nnU-Net~\cite{isensee2021nnu} & $0.811$ & $0.434$    \\

U-Net~\cite{jaeger2020retina} & $0.828$ & $0.459$    \\

\hline

Stage 1 (RadHop) & $0.815$ & $0.407$    \\

RadHop + RadHop-Net & $0.826$ & $0.422$ \\

\textbf{RadHop + RadHop-Net + wrMSE} & $\textbf{0.839}$ & $\textbf{0.468}$  \\  

\hline
\end{tabular}
\label{tab:benchmarking}
\end{center}
\end{table}

\vspace{-5mm}

\begin{figure}[htbp]
\centerline{\includegraphics[width=1.0\linewidth]{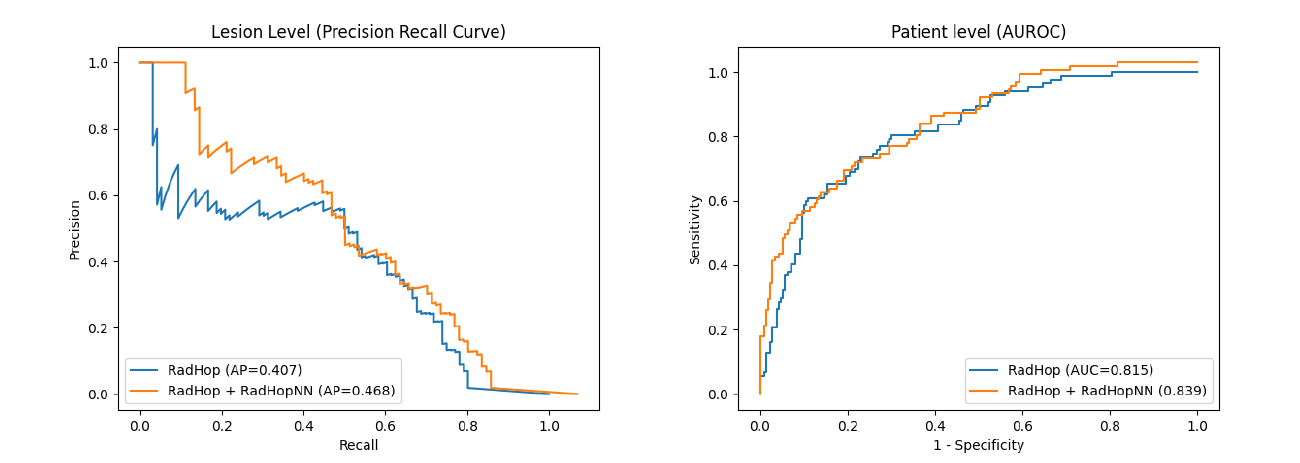}}
\vspace{-2mm}
\caption{Quantitative comparison of the detection performance before and after adding the RadHop-Net.}
\label{fig:quantitative}
\end{figure}

We evaluate the RadHop-Net pipeline for performance benchmarking using the patient and lesion-level performance metrics. The Area Under the Receiver Operating Curve (AUROC) between the sensitivity and false positive rate curve is used for the patient level metric. In contrast, the Average Precision (AP) is used for the lesion level predictions.

For quantitative analysis, we compare our proposed pipeline against the U-Net~\cite{jaeger2020retina} and nnU-Net~\cite{isensee2021nnu} baseline methods, popular for medical image analysis tasks. Looking at Table~\ref{tab:benchmarking}, we can infer that the proposed pipeline achieves a competitive performance with other DL-based approaches. Stage 1 alone achieves a performance of $0.815$ AUROC at the patient level and $0.407$ AP at the lesion level. Adding Stage 2 and RadHop-Net, one can see how the AP performance significantly improves to $0.468$, boosting the AUROC to $0.839$ (see Fig.~\ref{fig:quantitative}). Moreover, the proposed wrMSE is conducive to achieving this improvement, increasing the AP results by $+0.061$, over the unweighted MSE metric.

\begin{figure}[htbp]
\centerline{\includegraphics[width=1.0\linewidth]{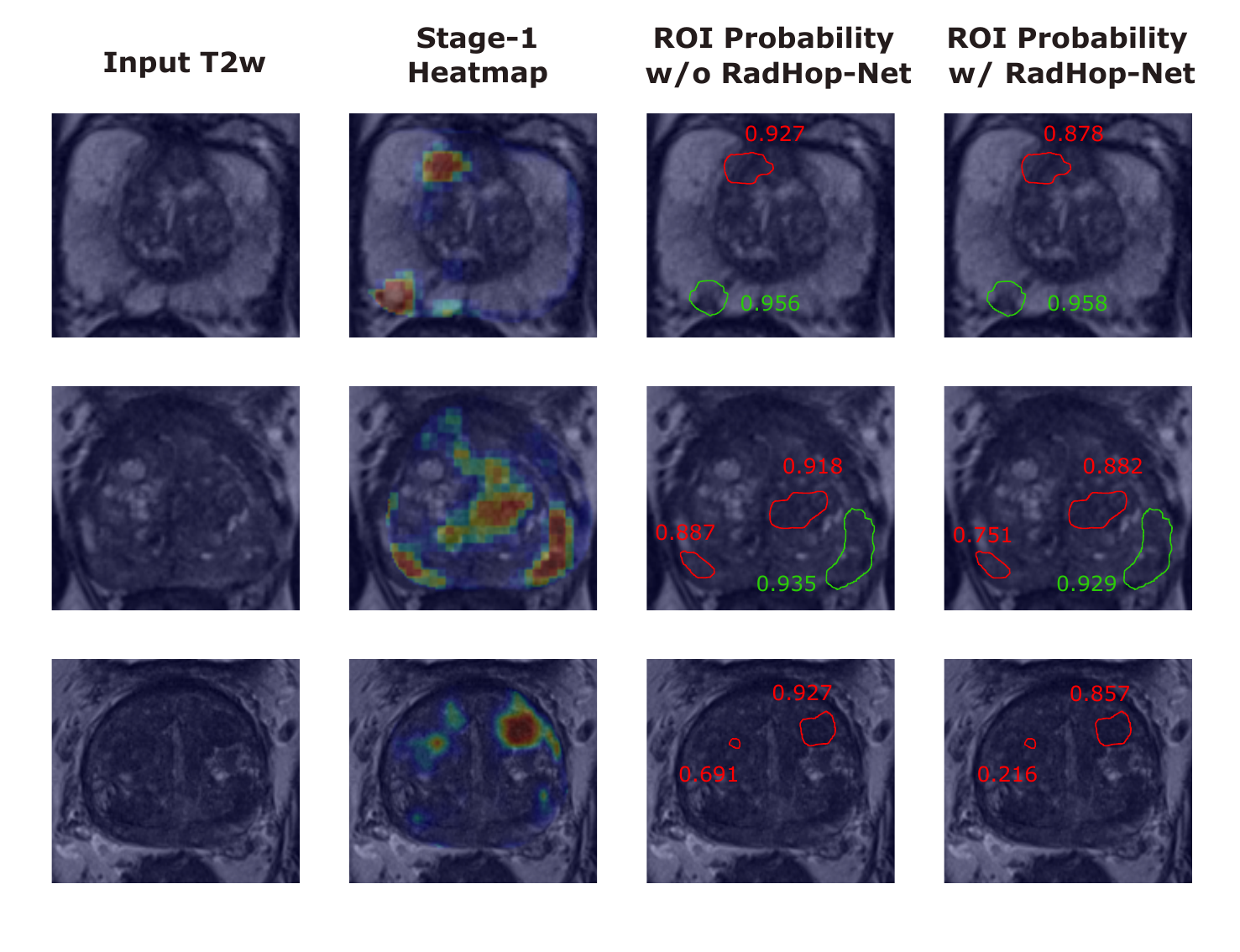}}
\vspace{-5mm}
\caption{Three examples from the pi-cai dataset, demonstrating the effectiveness of the RadHop-Net. False positives are shown in red, and true positives in green.}
\label{fig:qualitative}
\end{figure}

After RadHop-Net compensates for the Stage 1 ROI probability error residue, the average FP's probability decreases, while the actual csPCa lesions probability remains high (Fig.~\ref{fig:qualitative}). On top of its efficacy, RadHop-Net maintains a small model size with $54,585$ trainable weights, while most CNN models for similar tasks entail millions of parameters.

The experiments that were conducted display the effectiveness of the proposed RadHop-Net module to reduce the FP rate. Also, the proposed wrMSE loss is crucial to increase further the lesion level performance. It is also demonstrated that the RadHop feature space can provide a solid basis for data-driven radiomics and helps to reduce the complexity of RadHop-Net.

\vspace{-2mm}

\section{Conclusion}

This paper presents a novel pipeline that decouples from end-to-end solutions. It adopts a two-stage solution based on the RadHop data-driven radiomics. This unique approach proposes RadHop-Net, a custom CNN that receives radiomics as input and predicts a probability residue to compensate for Stage 1 errors. Also, a novel loss function for regression problems was introduced to balance the regression values between FPs and TPs. The overall pipeline achieves a high performance while maintaining a significantly lower complexity than other state-of-the-art works.


\begin{thebibliography}{00}
\bibitem{siegelstats} Siegel, R.L., Miller, K.D., Wagle, N.S., Jemal, A., 2023. Cancer statistics, 2023. CA: a cancer journal for clinicians 73, 17–48.


\bibitem{galey_psa} J. Galey, L., Olanrewaju, A., Nabi, H., Paquette, J. S., Pouliot, F., \& Audet-Walsh, É. (2024). PSA, an outdated biomarker for prostate cancer: In search of a more specific biomarker, citrate takes the spotlight. The Journal of Steroid Biochemistry and Molecular Biology, 243, 106588.


\bibitem{seo_pirads} Seo, J.W., Shin, S.J., Taik Oh, Y., Jung, D.C., Cho, N.H., Choi, Y.D., Park, S.Y., 2017. Pi-rads version 2: detection of clinically significant cancer in patients with biopsy Gleason score 6 prostate cancer. American Journal of Roentgenology 209, W1–W9.


\bibitem{algohary_radiomics} Algohary, A.,Viswanath, S., Shiradkar, R., Ghose, S., Pahwa, S., Moses, D., Jambor, I., Shnier, R., Böhm, M., Haynes, A.M., et al., 2018. Radiomic features on MRI enable risk categorization of prostate cancer patients on active surveillance: Preliminary findings. Journal of Magnetic Resonance Imaging 48, 818–828.

\bibitem{filos_radiomics} Filos, D., Fotopoulos, D., Rouni, M. A., \& Chouvarda, I. (2024, May). Machine Learning-Based Whole Gland Radiomics Analysis for Prostate Cancer Classification. In 2024 IEEE International Symposium on Biomedical Imaging (ISBI) (pp. 1-5). IEEE.


\bibitem{Yu_contextual} Yu, X., Lou, B., Shi, B., Winkel, D., Arrahmane, N., Diallo, M., Meng, T., von Busch, H., Grimm, R., Kiefer, B., et al., 2020a. False positive reduction using multiscale contextual features for prostate cancer detection in multi-parametric MRI scans, in 2020 IEEE 17th international
Symposium on biomedical imaging (ISBI), IEEE. pp. 1355–1359.

\bibitem{magoul_RadHop} Magoulianitis, V., Yang, J., Yang, Y., Xue, J., Kaneko, M., Cacciamani, G., ... \& Nikias, C. (2024). PCa-RadHop: A transparent and lightweight feed-forward method for clinically significant prostate cancer segmentation. Computerized Medical Imaging and Graphics, 102408.

\bibitem{Saab} Liu, X., Xing, F., Gaggin, H. K., Wang, W., Kuo, C. C. J., El Fakhri, G., \& Woo, J. (2021, November). Segmentation of cardiac structures via successive subspace learning with saab transform from cine MRI. In 2021 43rd Annual International Conference of the IEEE Engineering in Medicine \& Biology Society (EMBC) (pp. 3535-3538). IEEE.

\bibitem{Saha_fpreduction} Saha, A., Hosseinzadeh, M., \& Huisman, H. (2021). End-to-end prostate cancer detection in bpMRI via 3D CNNs: effects of attention mechanisms, clinical priori and decoupled false positive reduction. Medical image analysis, 73, 102155.

\bibitem{Zhang_crossmodal} Zhang, G., Shen, X., Zhang, Y. D., Luo, Y., Luo, J., Zhu, D., ... \& Lu, J. (2021). Cross-modal prostate cancer segmentation via self-attention distillation. IEEE Journal of Biomedical and Health Informatics, 26(11), 5298-5309.

\bibitem{Wang_LNT} Wang, X., Mishra, V. K., \& Kuo, C. C. J. (2023, December). Enhancing Edge Intelligence with Highly Discriminant LNT Features. In 2023 IEEE International Conference on Big Data (BigData) (pp. 3880-3887). IEEE.

\bibitem{Vente_Unet} De Vente, C., Vos, P., Hosseinzadeh, M., Pluim, J., \& Veta, M. (2020). Deep learning regression for prostate cancer detection and grading in bi-parametric MRI. IEEE Transactions on Biomedical Engineering, 68(2), 374-383.

\bibitem{Huang_Unet} Huang, X., Zhang, B., Zhang, X., Tang, M., Miao, Q., Li, T., \& Jia, G. (2021). Application of U-Net based multiparameter magnetic resonance image fusion in the diagnosis of prostate cancer. IEEE Access, 9, 33756-33768.

\bibitem{isensee2021nnu} Isensee, F., Jaeger, P.F., Kohl, S.A., Petersen, J., Maier-Hein, K.H., 2021. Nnu-net: a self-configuring method for deep learning-based biomedical image segmentation. Nature Methods 18, 203–211

\bibitem{jaeger2020retina} Jaeger, P.F., Kohl, S.A., Bickelhaupt, S., Isensee, F., Kuder, T.A., Schlemmer, H.P., Maier-Hein, K.H., 2020. Retina u-net: Embarrassingly simple
exploitation of segmentation supervision for medical object detection, in Machine Learning for Health Workshop, PMLR. pp. 171–183.

\bibitem{picai} Saha, A., Bosma, J., Twilt, J., van Ginneken, B., Yakar, D., Elschot, M., Veltman, J., Fütterer, J., de Rooij, M., et al., 2023. Artificial intelligence and radiologists at prostate cancer detection in mri—the pi-cai challenge, in: Medical Imaging with Deep Learning, short paper track.

\bibitem{b7} M. Young, The Technical Writer's Handbook. Mill Valley, CA: University Science, 1989.
\end{thebibliography}
\end{document}